\def\year{2016}\relax
\documentclass[letterpaper]{article}
\usepackage{aaai16}
\usepackage{times}
\usepackage{helvet}
\usepackage{courier}
\usepackage{amsmath,amssymb,amsthm}

\usepackage{multicol}

\usepackage{graphicx}
\usepackage{color}
\usepackage{subfigure} 

\usepackage{mathtools}
\DeclarePairedDelimiter{\ceil}{\lceil}{\rceil}

\usepackage{algorithm}
\usepackage{algorithmic}

\usepackage{multirow}

\usepackage{url}

\graphicspath{{images/}} % Directory in which figures are stored

\makeatletter

\newcommand{\Rmnum}[1]{\expandafter\@slowromancap\romannumeral #1@}
\makeatother

\newtheorem{assumption}{Assumption}
\newtheorem{theorem}{Theorem}
\newtheorem{corollary}{Corollary}
\newtheorem{prop}{Proposition}
\newtheorem{lemma}{Lemma}

\def\R{\mathbb{R}}

\def\tx{\tilde{x}}
\def\g{\gamma}
\def\tf{\tilde{f}}
\def\tg{\tilde{g}}

\def\tP{\tilde{P}}
\def\mU{\mathcal{U}}
\def\mM{\mathcal{M}}
\def\CE{\mathbb{E}}

\begin{document}
\title{Fast Nonsmooth 
Regularized Risk 
Minimization with Continuation\\}
\author{Shuai Zheng  \hspace{.2in} Ruiliang Zhang \hspace{.2in} James T. Kwok\\
Department of Computer Science and Engineering\\
Hong Kong University of Science and Technology\\
Hong Kong\\
\{szhengac, rzhangaf, jamesk\}@cse.ust.hk\\
}
\maketitle
\begin{abstract}
In regularized risk minimization, the associated optimization problem becomes particularly difficult when
both the loss and regularizer are nonsmooth. Existing approaches either have slow or unclear
convergence properties, are restricted to limited problem subclasses, or require careful setting of a
smoothing parameter. In this paper, we propose a continuation algorithm that is applicable to
a large class of 
nonsmooth regularized risk minimization problems,
can be flexibly used with a number of existing 
solvers for the underlying smoothed subproblem, and with convergence results on the whole
algorithm rather than just one of its subproblems.  In particular, when accelerated solvers
are used, the proposed algorithm achieves the fastest known rates of
$O(1/T^2)$ on strongly convex problems, and $O(1/T)$ on general convex problems. Experiments on nonsmooth
classification and regression tasks demonstrate that the proposed algorithm outperforms the state-of-the-art.
\end{abstract}

\section{Introduction}

In regularized risk minimization, one has to minimize the sum of an empirical loss and a regularizer.
When both are smooth, 
it can be easily 
optimized
by a variety of solvers
\cite{nesterov2004introductory}. 
In particular, 
a popular choice
for big data applications
is stochastic gradient descent (SGD), which is easy to implement and highly scalable \cite{kushner2003stochastic}.
For many nonsmooth regularizers (such as the
$\ell_1$ and nuclear norm regularizers),
the corresponding regularized risks can  still be efficiently 
minimized by the proximal gradient algorithm and its accelerated variants \cite{nesterov2013gradient}.
However, when the regularizer is smooth but the loss is nonsmooth
(e.g., the hinge loss and absolute loss),
or when both
the loss and regularizer
are nonsmooth,
proximal gradient algorithms are not directly applicable.

On nonsmooth problems, SGD can still be used,
by simply replacing the gradient with  subgradient.
However, the information contained in the subgradient 
is much less informative
\cite{nemirovsky1983problem}, and 
convergence is
hindered. 
On general convex problems, SGD converges at a rate of
$O(\log T/\sqrt{T})$,
where $T$ is the number of iterations; whereas
on strongly convex problems, the rate is
$O(\log T/T)$.
In contrast,
its smooth counterparts  converge with the much faster
$O(1/\sqrt{T})$ and $O(1/T)$ rates, respectively
\cite{rakhlin2012making,shamir2013stochastic}.
Recently,
\citeauthor{shamir2013stochastic} (\citeyear{shamir2013stochastic})
recovered these rates by using a polynomial-decay averaging scheme on the SGD
iterates.
However, a major drawback is that it does not exploit properties of the regularizer. For
example, when used with a 
sparsity-inducing
regularizer, its solution obtained may not be
sparse
\cite{duchi2009efficient}.

\citeauthor{nesterov2005smooth} (\citeyear{nesterov2005smooth}) proposed 
to smooth the nonsmooth objective so that it can then be efficiently optimized.
This smoothing approach is now popularly used for nonsmooth optimization. 
However, the optimal smoothness parameter 
needs to be known in advance.
This restriction is later avoided by the (batch) excessive gap algorithm \cite{nesterov2005excessive}. 
In the stochastic setting,
\citeauthor{ouyang2012stochastic} (\citeyear{ouyang2012stochastic})
combined Nesterov's smoothing with SGD.
Though these methods achieve the fastest known
convergence rates in the batch and stochastic
settings respectively, 
they assume a Lipschitz-smooth regularizer,
and nonsmooth regularizers (such as the sparsity-inducing regularizers) cannot  be used.

Recently, based on the observation that the training set is indeed finite, a number of fast 
stochastic algorithms are proposed for both smooth and composite optimization problems
\cite{schmidt2013minimizing,johnson2013accelerating,xiao2014proximal,Mairal2013,defazio-14}. 
They are based on the idea of variance reduction, and attain comparable
convergence rates as their batch counterparts.
However, they are not applicable when both the loss and regularizer are nonsmooth. 
To alleviate this,
\citeauthor{shalev2014accelerated} 
(\citeyear{shalev2014accelerated}) suggested running these 
algorithms on the smoothed approximation obtained by Nesterov's smoothing. However, as in \cite{nesterov2005smooth}, 
it requires a careful setting of the
smoothness parameter. 
Over-smoothing deteriorates solution quality, while under-smoothing
slows down convergence.

The problem of setting the smoothness parameter can be alleviated by continuation \cite{becker2011nesta}.
It solves a sequence of smoothed problems, in which the smoothing parameter is 
gradually reduced 
from a large value (and the corresponding smoothed problem is easy to solve) to a small value
(which leads to a solution closer to that of the original nonsmooth problem). Moreover,
solution of the intermediate problem is used to warm-start the next smoothed problem. 
This approach is also similar to that of gradually changing the regularization parameter in \cite{hale2007fixed,wen2010fast,mazumder2010spectral}. 
Empirically, continuation converges much faster than the use of a fixed smoothing parameter
\cite{becker2011nesta}. 
However, the theoretical convergence rate obtained  in
\cite{becker2011nesta}
is only for one 
stage  of the continuation algorithm 
(i.e.,  on the smoothed problem with a particular smoothing parameter), while the convergence
properties for the whole algorithm are not clear. 
Recently, 
\citeauthor{xiao2012proximal} (\citeyear{xiao2012proximal}) obtained a linear convergence
rate for their continuation algorithm, though 
only 
for the special case of  $\ell_1$-regularized least squares regression.

In this paper, we consider the general nonsmooth optimization setting, in which 
both the loss and regularizer may be nonsmooth.
The proposed continuation algorithm can be flexibly used 
with a variety of existing batch/stochastic solvers 
in each stage.  Theoretical analysis 
shows that  
the proposed algorithm, with this wide class of solvers, achieves the rate of
$O(1/T^2)$ on strongly convex problems, and $O(1/T)$ on general convex problems. 
These are the fastest known rates for nonsmooth optimization.
Note that these rates are for the whole algorithm, not just one of its stages as in
\cite{becker2011nesta}. 
Experiments on nonsmooth classification and regression models demonstrate that the
proposed algorithm outperforms the state-of-the-art.

\noindent
{\bf Notation}.
For $x,y\in\R^d$, $\|x\|_2=\sqrt{\sum_{i=1}^d x_i^2}$ is its $\ell_2$-norm, 
$\|x\|_1 = \sum_{i=1}^d|x_i|$ is its $\ell_1$-norm, and $\langle x, y\rangle$ is the dot product between $x, y$. Moreover,
$\partial f$ denotes the subdifferential of a nonsmooth function $f$, if $f$ is differentiable, then $\nabla f$ denotes its gradient. 
$I$ is the identity matrix.

%%%%%%%%%%%%%%%%%%%%%%%%%%%%%%%%%%%%%%%%%%%%%%%%%%%%%%%%%%%%%%%%%%%%%%%%%%%%%%%%%%%%

\section{Related Work}

Consider nonsmooth functions of the form 
\begin{eqnarray} \label{eq:dual}
g(x) = \hat{g}(x) + \max_{u \in \mU}[\langle Ax,u \rangle - Q(u)],
\end{eqnarray}
where 
$\hat{g}$ is convex, continuously differentiable with $\hat{L}$-Lipschitz-continuous
gradient,
$\mU \subseteq \R^p$ is convex, $A \in \R^{p \times d}$, 
and $Q$ is a continuous convex function.
\citeauthor{nesterov2005smooth}
(\citeyear{nesterov2005smooth})
proposed the following smooth approximation:
\begin{eqnarray} \label{eq:smoothed}
\tg_\g(x) = \hat{g}(x) + \max_{u \in \mU} \left[\langle Ax,u \rangle - Q(u) - \gamma \omega(u)\right],
\end{eqnarray}
where 
$\gamma$ is a smoothness parameter,
and 
$\omega$ is a nonnegative $\zeta$-strongly convex function.

For example,  consider
the hinge loss $g(x) = \max(0, 1 - y_iz_i^Tx)$, where $x$ is the 
linear model
parameter, and $(z_i,y_i)$ is the $i$th training sample with $y_i \in \{\pm 1\}$.
Using $\omega(u) = \frac{1}{2}\|u\|_2^2$, 
$g$
can be smoothed 
to 
\cite{ouyang2012stochastic}
\begin{equation} \label{eq:smoothed_hinge}
\tg_\g(x) = \left\{ \begin{array}{ll}
0 & y_iz_i^Tx \geq 1 \\
1 - y_iz_i^Tx - \frac{\gamma}{2} & y_iz_i^Tx < 1 - \gamma \\
\frac{1}{2\gamma}(1 - y_iz_i^Tx)^2 & \text{otherwise}
\end{array} \right..
\end{equation} 
Similarly, the $\ell_1$ loss $g(x) = |y_i - z_i^Tx|$ can be smoothed to
\begin{equation} \label{eq:smoothed_absolute}
\tg_\g(x) = \left\{ \begin{array}{ll}
y_i - z_i^Tx - \frac{\gamma}{2} & y_i - z_i^Tx \geq \gamma \\
-(y_i - z_i^Tx) - \frac{\gamma}{2} & y_i - z_i^Tx < - \gamma \\
\frac{1}{2\gamma}(y_i - z_i^Tx)^2 & \text{otherwise}
\end{array} \right..
\end{equation} 
Other examples in machine learning include popular regularizers such as the 
$\ell_1$,
total variation \cite{becker2011nesta},
overlapping group lasso,
and graph-guided fused lasso \cite{chen-12}.

Minimization of the smooth (and convex) $\tg_\gamma$
can be performed efficiently using
first-order methods,
including the so-called ``optimal method'' and its variants
\cite{nesterov2005smooth}
that achieve the optimal convergence rate.

%%%%%%%%%%%%%%%%%%%%%%%%%%%%%%%%%%%%%%%%%%%%%%%%%%%%%%%%%%%%%%%%%%%%%%%%%%%%%%%%%%%%

\section{Nesterov Smoothing with Continuation}

\begin{table*}[t]
\caption{Examples of non-accelerated and accelerated solvers.
Note that Prox-GD and APG are batch solvers while the others are stochastic solvers.
Here, $\theta$ and $p$ are
parameters related to the stepsize, and are fixed across stages. In particular, $\theta \in (0, 0.25)$ and satisfies $(1 - 4\theta)\rho_s - 4\theta > 0$, and $p \in (0, 1)$ and satisfies $\rho_s > \frac{p(2+p)}{1-p}$.
Accelerated Prox-SVRG has $T_s = O(\sqrt{\kappa_s} \log(1 / \rho_s))$ only when a
sufficiently large mini-batch is used. }
\label{alg_detail}
\begin{center}
\scalebox{0.9}{
\begin{tabular}{cc|c|c|c|c|c}
\hline
\multicolumn{2}{c|}{} & $T_s$ & $\phi(\rho_s)$ & $a$ & $b$ & $c$\\ \hline
\multirow{4}{*}{non-accelerated}& Prox-GD \cite{nesterov2013gradient} & $4\kappa_s\log(1/\rho_s)$ & $\log(1/\rho_s)$ & $4$ & $0$ & $0$\\ 
& Prox-SVRG \cite{xiao2014proximal} & $\frac{\theta}{(1 - 4\theta)\rho_s - 4\theta}\left(\kappa_s + 4\right)$ & $\frac{1}{(1 - 4\theta)\rho_s - 4\theta}$ & $\theta$ & $4\theta$ & $0$\\ 
& SAGA \cite{defazio-14} & $\frac{3n}{\rho_s}\left(\frac{3\kappa_s}{n} + 1\right)$ &$\frac{1}{\rho_s}$ 
& $9$ & $3n$ & $0$\\ 
& MISO \cite{Mairal2013} & $\frac{n\kappa_s}{\rho_s}$ & $\frac{1}{\rho_s}$ & $n$ & $0$ & $0$\\ \hline
\multirow{2}{*}{accelerated} & APG \cite{schmidt2011convergence} & $\sqrt{\kappa_s}\log(2/\rho_s)$ & $\log(2/\rho_s)$ & $1$ & $0$ & $0$\\ 
& Accelerated Prox-SVRG \cite{nitanda2014stochastic} &
$\sqrt{\kappa_s}\frac{\sqrt{2}}{(1 - p)}\log \left(\frac{1}{\frac{\rho_s}{2+p} -
\frac{p}{1-p}} \right)$ & $\log \left(\frac{1}{\frac{\rho_s}{2+p} - \frac{p}{1-p}}
\right)$ & $\frac{\sqrt{2}}{(1 - p)}$ & $0$ & $0$\\ \hline
\end{tabular}
}
%}
\end{center}
\end{table*}

%In this paper, we 
Consider the following nonsmooth minimization problem
\begin{eqnarray} \label{eq:problem}
\min_x P(x) \equiv f(x) + r(x),
\end{eqnarray}
where both $f$  and $r$ are convex and nonsmooth. In machine learning, $x$ usually corresponds to the model parameter,
$f$ is the loss, and $r$ the regularizer.  We assume that the loss $f$ on a set of $n$ training samples
can be decomposed as $f(x) =
\frac{1}{n}\sum_{i=1}^nf_i(x)$, where $f_i$ is the loss value on the $i$th sample.
Moreover, each $f_i$ can be written as
in (\ref{eq:dual}), i.e.,
$f_i(x) = \hat{f}_i(x) + \max_{u \in \mU}\left[\langle A_ix,u \rangle -
Q(u)\right]$. 
One can then apply Nesterov's smoothing,
and $P(x)$ in (\ref{eq:problem}) is smoothed to
\begin{eqnarray} \label{eq:smoothed_obj}
\tP(x) = \tf_\g(x) + r(x),
\end{eqnarray}
where $\tf_\g(x) = \frac{1}{n}\sum_{i=1}^n\tf_i(x)$ and
\begin{equation} \label{eq:tfi}
\tf_i(x)=
\hat{f}_i(x) + \max_{u \in \mU} \left[\langle A_ix,u \rangle - Q(u) - \gamma
\omega(u)\right].
\end{equation} 
As for $r$, we assume that it is ``simple", namely that
its proximal operator, $\text{prox}_{\lambda r}(\cdot) \equiv \arg\min_x \frac{1}{2} \|x -
\cdot\|^2 + \lambda r(x)$ for any $\lambda>0$,
can be easily computed \cite{parikh2014proximal}.

%%%%%%%%%%%%%%%%%%%%%%%%%%%%%%%%%%%%%%%%%%

\subsection{Strongly Convex Objectives}

In this section, we  assume that $P$ is $\mu$-strongly convex. This 
strong convexity may come from $f$ (e.g., $\ell_2$-regularized hinge loss)
or $r$ (e.g., elastic-net regularizer)
or both.

\begin{assumption} \label{assum_strong}
$P$ is $\mu$-strongly convex, i.e., there exists $\mu > 0$
such that 
$P(y) \geq P(x) + \xi^T(y - x) + \frac{\mu}{2}\|y - x\|^2_2, \forall \xi \in \partial P(x)$
and $x, y \in \R^d$.
\end{assumption}

The proposed
algorithm 
is based on continuation. It
proceeds in stages, and a smoothed problem
is solved in each stage \cite{becker2011nesta}.
The smoothness parameter 
is gradually reduced across stages,
so that the smoothed problem becomes closer and closer to the original one.
In each stage, an iterative solver
$\mM$ is used to solve the
smoothed problem.
It returns an approximate solution, which is then used to warm-start
the next stage.

In stage $s$, let the smoothness parameter be $\gamma_s$,
the smoothed objective in (\ref{eq:smoothed_obj}) be
$\tP_s(x)$, 
$x_s^*=\arg\min_x \tP_s(x)$, and
$\tx_s$ be the solution returned by $\mM$. 
As $\mM$ is warm-started by $\tx_{s-1}$,
the error  
before running $\mM$ 
is $\tP_{s}(\tx_{s-1}) - \tP_{s}(x_s^*)$.
At the end of stage $s$, 
we assume that 
the error is reduced by a factor of 
$\rho_s$.
The expectation $\CE$ below is over the stochastic choice of training samples
for a stochastic solver.
For a deterministic solver, this expectation can be dropped.  

\begin{assumption} \label{assum_linear}
$\CE\tP_s(\tx_{s}) - \tP_s(x_s^*) \leq \rho_s(\tP_{s}(\tx_{s-1}) - \tP_{s}(x_s^*))$, where 
$\rho_s \in (0, 1)$.
\end{assumption}

We consider two types of solvers, which differ in the number of iterations ($T_s$) it takes to satisfy
Assumption~\ref{assum_linear}.
\begin{enumerate}
\item Non-accelerated solvers:
$T_s=
a\kappa_s\phi(\rho_s) + b\phi(\rho_s) + c$;
\item Accelerated solvers:
$T_s=a\sqrt{\kappa_s}\phi(\rho_s) + b\phi(\rho_s) + c$.
\end{enumerate}
Here, 
$\kappa_s$ 
is the condition number 
of the objective,
$a, b, c\geq 0$ are constants
not related to $\kappa_s$ and $\phi(\rho_s)$. Moreover,
$\phi$ satisfies (i) $\phi(\rho_s) > 0$ and
non-increasing 
for $\rho_s \in (0, 1)$; (ii) $\phi(\rho_s)$ is not related to $\kappa_s$. 
Note that when $\kappa_s$ is large
(as is typical when the smoothed problem approaches the original problem),  
non-accelerated solvers need a larger $T_s$ than accelerated solvers. 
Table~\ref{alg_detail} shows some non-accelerated and accelerated solvers 
popularly used in machine learning.

Algorithm~\ref{alg:nsmm}
shows
the proposed procedure, which will be called CNS (Continuation for NonSmooth optimization).
It is similar to  that in
\cite{becker2011nesta},
which however does not have convergence results.
Moreover, a small but important difference is that 
Algorithm~\ref{alg:nsmm} specifies how 
$T_s$ should be updated across stages, and this is essential for proving convergence.
Note the different update options 
for non-accelerated and accelerated solvers.

\begin{algorithm}[htbp]
\caption{CNS algorithm for strongly convex problems.}
\label{alg:nsmm}
\begin{algorithmic}[1]
\STATE {\bfseries Input:} 
number of iterations $T_1$ and
smoothness parameter $\gamma_1$
for stage 1,
and shrinking parameter $\tau > 1$.
   \STATE {\bfseries Initialize:} $\tx_0$.
   \FOR{$s=1, 2, \dots$}
   \STATE{$\tP_s \leftarrow \text{smooth $P$ with smoothing parameter $\gamma_s$}$};
   \STATE{$\tx_s \leftarrow$ minimize $\tP_s(x)$ by running $\mM$ for $T_s$ iterations};
   \STATE{$\gamma_{s+1} = \gamma_s / \tau$};
   \STATE{Option $\mathrm{\Rmnum{1}}$
	(non-accelerated solvers): $T_{s+1} = \tau T_s$}; 
   \STATE{Option $\mathrm{\Rmnum{2}}$
	(accelerated solvers): $T_{s+1} = \sqrt{\tau} T_s$}; 
   \ENDFOR
   \STATE {\bfseries Output:} $\tx_{s}.$
\end{algorithmic}
\end{algorithm}

The following Lemma shows that when $T_1$ is large enough, 
error reduction  
can be guaranteed 
across all stages.
\begin{lemma} \label{lemma_rho}
For both non-accelerated and accelerated solvers, if $T_1$ is large enough such
that $\rho_1 \leq 1/\tau^2$, then $\rho_s \leq 1/\tau^2$ for all $s > 1$.
\end{lemma}

If $\kappa_1$ is known, a sufficiently large $T_1$ can be obtained from Table~\ref{alg_detail};
otherwise, we can obtain $T_1$ by ensuring $\tP_1(\tx_1) \leq \tP_1(\tx_0)/\tau^2 $, which
then implies $\tP_1(\tx_1) -  \tP_1(x_1^*) \leq (\tP_1(\tx_0) - \tP_1(x_1^*))/\tau^2$.

%%%%%%%%%%%%%%%%%%%%%%%%%%%%%%%%%%%%%%%%%%

\subsubsection{Convergence when Non-Accelerated Solver is used}

Let $x^*=\arg\min_x P(x)$, and $D_u = \max_{u \in \mU} \omega(u)$.
The following Lemma shows that if $x$ is an $\epsilon$-accurate
solution
of the $\tP_s$ (i.e.,
$\tP_s(x) - \tP_s(x_s^*)  \leq \epsilon$), it is also an $(\epsilon +
\gamma_sD_u)$-accurate solution of the original objective $P$.

\begin{lemma} \label{lemma_smooth_gap}
$\tP_s(x) - \tP_s(x_s^*) - \gamma_s D_u \leq P(x) - P(x^*) \leq \tP_s(x) - \tP_s(x_s^*) + \gamma_s D_u$.
\end{lemma}
Since Lemma~\ref{lemma_smooth_gap} holds for any $x$, it also holds in expectation, i.e., $\CE\tP_s(\tx_s) - \tP_s(x_s^*) - \gamma_s D_u \leq \CE P(\tx_s) - P(x^*) \leq \CE\tP_s(\tx_s) - \tP_s(x_s^*) + \gamma_s D_u$.

\begin{theorem} \label{theorem:non-acc}
Assume that $T_1$ 
in Algorithm~\ref{alg:nsmm}
is large enough
so that $\rho_1 \leq 1/\tau^2$. 
When non-accelerated solvers are used,
\begin{eqnarray} \label{eq:rate_non_acc}
\!\!\!\!  \CE P(\tx_S) \!\! -\!\!  P(x^*) & \!\!\!\!\! \leq \!\!\!\!\! & \!\!
\left(\prod_{s = 1}^S \rho_s \!\!\! \right)\!\! \left(P(\tx_{0}) \!\! - \!\! P(x^*)
\right) \!\! +\!\!  O\!\!\left(\!\! \frac{\gamma_1D_u}{T}\!\!\! \right)\!\!,
\end{eqnarray}
where
$S$ is the number of stages,
$T = \sum_{s=1}^S T_s$, and $\prod_{s = 1}^S \rho_s = O(1/T^2)$.
\end{theorem}
The first term on the RHS of (\ref{eq:rate_non_acc}) reflects the cumulative decrease of the objective 
after $S$ stages, while
the second term is due to smoothing. 
The condition $\rho_1 \leq 1/\tau^2$ is used to obtain the $O(1/T)$ rate in the last term
of (\ref{eq:rate_non_acc}).
If we instead require that $\rho_1 \leq 1/\tau$, it can be shown that the rate will be slowed to $O(\log T/T)$;
if $\rho_1 \leq 1/\sqrt{\tau}$, it degrades further to $O(1/\sqrt{T})$. On the other hand, if
$\rho_1 \leq 1/\tau^c$ with $c > 2$, the rate will not be improved. 

\begin{corollary} \label{coro_rate_non_acc_upper}
Together with Lemma~\ref{lemma_rho}, we have
\begin{eqnarray}  \label{eq:rate_non_acc_upper}
\!\!\!\!  \CE P(\tx_S) -P(x^*) & \!\!\!\!\! \leq \!\!\!\!\! & \frac{ P(\tx_{0}) - P(x^*)}{\tau^{2S}}
+O\left(\frac{\gamma_1D_u}{T}\right),
\end{eqnarray}
where $1/\tau^{2S} = O(1/T^2)$.
\end{corollary}

Existing stochastic algorithms such as SGD, FOBOS and RDA 
have a convergence rate of $O(\log T/T)$
\cite{rakhlin2012making,duchi2009efficient,xiao2009dual}, while here we have the faster
$O(1/T)$ rate. Recent works in
\cite{shamir2013stochastic,ouyang2012stochastic} also achieve a  $O(1/T)$ rate. 
However,
\citeauthor{shamir2013stochastic} 
(\citeyear{shamir2013stochastic})
use stochastic subgradient, and do not exploit 
properties of the regularizer
(such as sparsity). This can lead to 
inferior 
performance 
\cite{duchi2009efficient,xiao2009dual,mazumder2010spectral}. On the other hand,
\cite{ouyang2012stochastic} is restricted to $r \equiv 0$ in (\ref{eq:problem}).

Next, we compare with the case where continuation is not used
(i.e., $\gamma_s$  is a constant).
Equivalently, this corresponds to setting
$\tau = 1$ 
in 
Algorithm~\ref{alg:nsmm}.
 
\begin{prop}  \label{prop_rate_fixed}
When continuation is not used, let
$\rho \in (0, 1)$ be the error
reduction factor at each stage, and $\gamma > 0$ be the fixed smoothing parameter.
When either an accelerated or non-accelerated solver is used, 
\begin{eqnarray} \label{rate_fixed}
\!\!\!\!\!\!\!\CE P(\tx_S) \!-\! P(x^*) \!\!\!\!\!\!
& \leq & \!\!\!\!\!\! \rho^S(P(\tx_{0}) \!-\! P(x^*)) \!+\! (1 \!+\! \rho^S)\gamma D_u.
\end{eqnarray}
\end{prop}

\begin{prop} \label{prop:nonacc}
Assume that the two terms on the RHS of (\ref{eq:rate_non_acc_upper}) and (\ref{rate_fixed}) 
are equal to $\alpha\epsilon$ and $(1-\alpha)\epsilon$, respectively, where $\alpha>0$
and $\epsilon>0$. Let $\rho_1 
=\rho = 1/\tau^2$ in 
(\ref{eq:rate_non_acc}) 
and
(\ref{rate_fixed}). 
Assume that 
Algorithm~\ref{alg:nsmm}
needs a total of $T$
iterations to obtain an $\epsilon$-accurate solution, while its
fixed-$\gamma_s$
variant 
takes $T'$ iterations.
Then,
\begin{eqnarray*} 
T \!\!\! & \!\! \geq \!\! & \!\!\! \frac{\tau^S - 1}{\tau - 1}\left(a\kappa_1\phi\left(\frac{1}{\tau^2}\right) \!+\!
b\phi\left(\frac{1}{\tau^2}\right) \!+\! c\right), \\
T' \!\!\! & \!\! \geq \!\! & \!\!\! S\left(a\left(\frac{\tau^{2S} + 1}{\tau^{S+1} +
\tau^{S}}\kappa_1 \!+\! C\right)\phi\left(\frac{1}{\tau^2}\right) \!+\! b\phi\left(\frac{1}{\tau^2}\right) \!+\! c\right),
\end{eqnarray*}
where $S \geq  \log \left(\frac{\alpha\epsilon}{\left(P(\tx_{0}) -
P(x^*)\right)}\right)/\log\left(\frac{1}{\tau^2}\right)$, $C = \left(1 - \frac{\tau^{2S} +
1}{\tau^{S+1} + \tau^{S}}\right)\frac{K}{\mu}$, and $K$ is a constant,
\end{prop}
$T$ and $T'$ are usually dominated by the $a\kappa_1\phi(1/\tau^2)$ term, and $T'$ is roughly $S$ times that of $T$. This is also consistent with 
empirical observations that continuation is much faster than fixed
smoothing \cite{becker2011nesta}. 

%%%%%%%%%%%%%%%%%%%%%%%%%%%%%%%%%%

\subsubsection{Convergence when Accelerated Solver is used}

\begin{theorem} \label{theorem:acc}
Assume that $T_1$ 
in Algorithm~\ref{alg:nsmm}
is large enough so that $\rho_1 \leq 1/\tau^2$.
When accelerated solvers are used,
\begin{eqnarray*} \label{rate_acc}
\!\!\!\!  \CE P(\tx_S) \!\! -\!\!  P(x^*\!) & \!\!\!\!\! \leq \!\!\!\!\! & \!\left(\!\prod_{s = 1}^S
\rho_s\!\!\right)\!\! \left(P(\tx_{0}) \!\! - \!\! P(x^*)\right) \!\! +\!\!
O\!\!\left(\frac{\gamma_1D_u}{T^2}\!\!\right)\!\!.
\end{eqnarray*}
where 
$T=\sum_{s=1}^S T_s$, and $\prod_{s = 1}^S \rho_s = O(1/T^4)$.
\end{theorem}
As the $\rho_s$'s for non-accelerated and accelerated solvers are different, the $\prod_{s = 1}^S \rho_s$ term here is different from that in Theorem~\ref{theorem:non-acc}.
Moreover, 
the last term is improved from $O(1/T)$ in Theorem~\ref{theorem:non-acc} to $O(1/T^2)$
with accelerated solvers.
This is also better than the rates
of existing stochastic algorithms
($O(\log T/T)$ in \cite{duchi2009efficient,xiao2009dual} and $O(1/T)$ in
\cite{rakhlin2012making,shamir2013stochastic,ouyang2012stochastic}).
Besides, the black-box lower bound of $O(1/T)$ for strongly convex problems
\cite{agarwal2009information} does not apply here, as we have additional assumptions that
the objective is of the form in (\ref{eq:dual}) and the  number of
training samples is
finite.
Though the (batch) excessive gap algorithm \cite{nesterov2005excessive} also has a $O(1/T^2)$ rate,
it is limited to $r\equiv 0$ in (\ref{eq:problem}). 

As in Proposition~\ref{prop:nonacc}, the following shows that
if continuation is not used, the algorithm is roughly $S$ times slower.
\begin{prop} \label{prop:nonacc2}
With the same assumptions in 
Proposition~\ref{prop:nonacc},
\begin{eqnarray*} 
T \!\!\! & \!\! \geq \!\! & \!\!\! \frac{\sqrt{\tau}^S - 1}{\sqrt{\tau} -
1}\left(a\sqrt{\kappa_1}\phi\left(\frac{1}{\tau^2}\right) \!+\! b\phi\left(\frac{1}{\tau^2}\right) \!+\! c\right), \\
T' \!\!\!& \!\! \geq \!\! & \!\!\! S\left(a\sqrt{\frac{\tau^{2S} + 1}{\tau^{S+1} +
\tau^{S}}\kappa_1 \!+\! C}\phi\left(\frac{1}{\tau^2}\right) \!+\! b\phi\left(\frac{1}{\tau^2}\right) \!+ \!c\right),
\end{eqnarray*}
where $S,C$ are as defined in
Proposition~\ref{prop:nonacc}.
\end{prop}

%%%%%%%%%%%%%%%%%%%%%%%%%%%%%%%%%%%%%%%%%%

\subsection{General Convex Objectives}

When $P$ is not strongly convex, we add to it a small $\ell_2$ term (with weight
$\lambda_s$).
We then gradually decrease 
$\gamma_s$ and 
$\lambda_s$ simultaneously to approach the original problem. The revised procedure is shown in 
Algorithm~\ref{alg:nsmm_nst}.

\begin{algorithm}[htbp]
\caption{CNS algorithm for general convex problems.}
   \label{alg:nsmm_nst}
\begin{algorithmic}[1]
   \STATE {\bfseries Input:} 
	number of iterations $T_1$, smoothness parameter $\gamma_1$ and
	strong convexity parameter $\lambda_1$
	for stage 1,
	and shrinking parameter $\tau > 1$.
   \STATE {\bfseries Initialize:} $\tx_0$.
   \FOR{$s=1, 2, \dots$}
   \STATE{$\tP_s \leftarrow \text{smooth $P$ with smoothing parameter $\gamma_s$}$};
   \STATE{$\tx_s \leftarrow$ minimize $\tP_s(x) + \frac{\lambda_s}{2}\|x\|_2^2$ by running
	$\mM$ for $T_s$ iterations};
   \STATE{$\gamma_{s+1} = \gamma_s / \tau$; $\lambda_{s+1} = \lambda_s / \tau$};
   \STATE{Option $\mathrm{\Rmnum{1}}$ (non-accelerated solvers): $T_{s+1} = \tau^2 T_s$};
   \STATE{Option $\mathrm{\Rmnum{2}}$ (accelerated solvers): $T_{s+1} = \tau T_s$};
   \ENDFOR
   \STATE {\bfseries Output:} $\tx_{s}.$
\end{algorithmic}
\end{algorithm}

We assume
that there exists $R > 0$ such that
$\|x^*\|_2 \leq R$,
and $\|x_s^*\|_2 \leq R$ for all $s$.
Define $H_s(x) = \tP_s(x) + \frac{\lambda_s}{2}\|x\|_2^2$,
and let $x_s^*=\arg\min_x H_s(x)$.
The following assumption 
is similar to that for strongly convex problems.
\begin{assumption} \label{assum_linear_ns}
$\CE H_s(\tx_{s}) - H_s(x_s^*) \leq \rho_s(H_{s}(\tx_{s-1}) - H_{s}(x_s^*))$, where 
$\rho_s \in (0, 1)$.
\end{assumption}

\begin{theorem} \label{theorem:non-acc-ns}
Assume that $T_1$ 
in Algorithm~\ref{alg:nsmm_nst}
is large enough so that $\rho_1 \leq 1/\tau^2$. When non-accelerated solvers
are used,
\begin{eqnarray} \label{rate_non_acc_ns}
\CE P(\tx_S) \!\! - \!\! P(x^*)
& \!\!\!\!\!\!\!\!\!\!\!\! \leq \!\!\!\!\!\!\!\!\!\!\!\! & \left(\prod_{s = 1}^S \rho_s\right)\!\! \left(P(\tx_{0}) \!\! - \!\! P(x^*) \!\! + \!\! \frac{\lambda_1}{2}\|\tx_{0}\|_2^2\right)  \nonumber\\ 
&& + O\left(\frac{\lambda_1R^2/2}{\sqrt{T}}\right) + O\left(\frac{\gamma_1D_u}{\sqrt{T}}\right),
\end{eqnarray}
where $\prod_{s = 1}^S \rho_s = O(\frac{1}{T})$. For accelerated solvers,
\begin{eqnarray} \label{rate_acc_ns}
\CE P(\tx_S) \!\! - \!\! P(x^*)
& \!\!\!\!\!\!\!\!\!\!\!\! \leq \!\!\!\!\!\!\!\!\!\!\!\! & \left(\prod_{s = 1}^S \rho_s\right)\!\! \left(P(\tx_{0}) \!\! - \!\! P(x^*) \!\! + \!\! \frac{\lambda_1}{2}\|\tx_{0}\|_2^2\right)  \nonumber\\ 
&& + O\left(\frac{\lambda_1R^2/2}{T}\right) + O\left(\frac{\gamma_1D_u}{T}\right), 
\end{eqnarray}
where 
$\prod_{s = 1}^S \rho_s = O(\frac{1}{T^2})$.
\end{theorem}

For non-accelerated solvers, the $O(1/\sqrt{T})$ convergence rate in (\ref{rate_non_acc_ns}) is
only as good as that obtained 
in
\cite{xiao2009dual,duchi2009efficient,ouyang2012stochastic,shamir2013stochastic}.
Hence, they will not be studied further in the sequel.
However, for accelerated solvers, the $O(1/T)$ convergence rate in
(\ref{rate_acc_ns}) is faster than the $O(1/\sqrt{T})$ rate in
\cite{xiao2009dual,duchi2009efficient,ouyang2012stochastic,shamir2013stochastic}
and the $O(\frac{1}{T^2} + \frac{\log T}{T})$ rate in \cite{orabona-12}. 
The $O(1/T)$ convergence rate is also obtained in
\cite{nesterov2005excessive,nesterov2005smooth}, but again only for $r\equiv 0$ in (\ref{eq:problem}).

When continuation is not used,
the following results are analogous to those obtained in the previous section.
\begin{prop}
Let $\tx_0 = 0$. When continuation is not used, 
let $\rho$
be the error reduction factor at each stage.
When either an accelerated or non-accelerated solver is used,
\begin{eqnarray} \label{rate_fixed_ns}
\CE P(\tx_S) - P(x^*)
& \leq & \rho^S(P(\tx_{0}) - P(x^*)) \nonumber \\
&& + (1 + \rho^S)\gamma D_u + \frac{\lambda}{2}R^2.
\end{eqnarray}
\end{prop}

\begin{prop} \label{prop:nonacc_ns}
Let $\tx_0 = 0$.  Suppose that the three terms on the RHS of 
(\ref{rate_fixed_ns}) 
are equal to $\alpha\epsilon$, $\beta\epsilon$ and $\zeta\epsilon$, respectively, where $\alpha, \beta, \zeta >0$ and
$\alpha + \beta + \zeta = 1$. Let $\rho_1 
=\rho = 1/\tau^2$ in 
(\ref{rate_acc_ns})  and
(\ref{rate_fixed_ns}). 
Assume that 
Algorithm~\ref{alg:nsmm_nst} (with accelerated solver)
needs a total of $T$
iterations to obtain an $\epsilon$-accurate solution, while its fixed-$\gamma_s$ variant
takes $T'$ iterations.
Then,
\begin{eqnarray*} 
T  \!\!& \!\! \geq \!\! & \!\! \frac{\tau^S - 1}{\tau -
1}\left(a\sqrt{\kappa_1}\phi\left(\frac{1}{\tau^2}\right) + b\phi\left(\frac{1}{\tau^2}\right) + c\right), \\
T'  \!\!& \!\! \geq \!\! & \!\! S\left(a\sqrt{\frac{\tau^{2S} + 1}{(\tau + 1)^2}\kappa_1 \!+\!
C}\phi\left(\frac{1}{\tau^2}\right) + b\phi\left(\frac{1}{\tau^2}\right) + c\right),
\end{eqnarray*}
where $S \geq  \log \left(\frac{\alpha\epsilon}{\left(P(\tx_{0}) -
 P(x^*)\right)}\right)/\log\left(\frac{1}{\tau^2}\right)$, $C = \left(1 - \frac{\tau^{2S} +
 1}{(\tau + 1)^2}\right) + \left(\frac{\tau^S}{1 + \tau} - \frac{\tau^{2S} + 1}{(\tau +
 1)^2}\right)\frac{K}{\lambda_1}$ and $K$ is a constant.
\end{prop}

A summary of the convergence results is shown in Table~\ref{rate_com}.  As can be seen, the
convergence rates of the proposed CNS algorithm match the fastest known rates in nonsmooth
optimization, but CNS is
less restrictive and 
can exploit the composite structure of the optimization problem.

\begin{table}[ht]
\caption{Comparison with the fastest known convergence rates for nonsmooth optimization problem
(\ref{eq:dual}). The fastest known batch solver is restricted to 
$r\equiv 0$, while 
the fastest known stochastic solver does not exploit properties of $r$.}
\vskip .1in
\label{rate_com}
\begin{center}
\begin{tabular}{c|c|c|c|c}
 \hline 
strongly & batch & stochastic &  \multicolumn{2}{c}{CNS (batch/stochastic)} \\ 
convex & solver & solver &  non-accel.& accel.\\ \hline \hline
yes & $1/T^2 $ & $1/T$ & $1/T$ & $1/T^2$ \\
 \hline 
no & $1/T $ & $1/\sqrt{T}$ & $1/\sqrt{T}$ & $1/T$\\
\hline
\end{tabular}
\end{center}
\end{table}

%%%%%%%%%%%%%%%%%%%%%%%%%%%%%%%%%%%%%%%%%%%%%%%%%%%%%%%%%%%%%%%%%%%%%%%%%%%%%%%%%%%%

\section{Experiments}

%%%%%%%%%%%%%%%%%%%%%%%%%%%%%%%%%%%%%%%%%%

Because of the lack of space, we only report results on 
two data sets 
(Table~\ref{data_detail})
from the LIBSVM archive: 
(i) the popularly used 
classification data set
{\em rcv1};
and (ii)
{\em YearPredictionMSD}, the largest regression data in the LIBSVM archive, and is a subset
of the Million Song data set.
We use the hinge loss
for classification,  and
$\ell_1$ loss 
for regression. Both can be smoothed 
using Nesterov's smoothing
(to (\ref{eq:smoothed_hinge}) and (\ref{eq:smoothed_absolute}), respectively).
As for the regularizer, we use the 
\begin{enumerate}
\item elastic-net regularizer $r(x) = \nu_1\|x\|_1 + \frac{\nu_2}{2}\|x\|_2^2$
\cite{zou2005regularization}, and
problem (\ref{eq:problem}) is strongly convex; and
\item $\ell_1$ regularizer $r(x) = \nu_1\|x\|_1$, and
(\ref{eq:problem}) is (general) convex.
\end{enumerate}
Here, $\nu_1, \nu_2$ are tuned by 5-fold cross-validation. Obviously, all losses and regularizers are convex but nonsmooth.
We use mini-batch for all methods. The mini-batch size $b$ is 50 for {\em rcv1},
and 100 for {\em YearPredictionMSD}. 

\begin{table}[ht]
\vskip -.1in
\caption{Data sets used in the experiments.}
\label{data_detail}
\begin{center}
\begin{tabular}{cccc}
\hline
& \#train & \#test  & \#features \\ \hline
{\em rcv1}  &  20,242 & 677,399 & 47,236 \\
{\em YearPredictionMSD}   &  463,715 & 51,630 & 90 \\
\hline
\end{tabular}
\end{center}
\end{table}

The following stochastic algorithms
are compared:
\begin{enumerate}
\item Forward-backward splitting (FOBOS) \cite{duchi2009efficient},
a standard baseline for nonsmooth stochastic 
composite optimization.
\item SGD with polynomial-decay averaging (Poly-SGD) \cite{shamir2013stochastic}, the state-of-art for nonsmooth optimization.

\item Regularized dual averaging (RDA) \cite{xiao2009dual}: This is another
state-of-the-art for sparse learning problems.

\item The proposed CNS algorithm: We use proximal SVRG (PSVRG) \cite{xiao2014proximal} as the underlying non-accelerated solver,
and accelerated proximal SVRG  (ACC-PSVRG) \cite{nitanda2014stochastic} as the accelerated solver.
The resultant procedures are denoted
CNS-NA and CNS-A, respectively.
We set 
$\gamma_1 = 0.01, \tau = 2$, and
$T_1 = \ceil{n/b}$. Empirically, this ensures $\rho_1 \leq 1/\tau^2$ (in
Theorems~\ref{theorem:non-acc} and \ref{theorem:non-acc-ns})
on the two data sets. 
\end{enumerate}
Note that 
FOBOS, RDA and the proposed CNS can effectively make use of the composite structure of the problem,
while 
Poly-SGD 
cannot.
For each method, the stepsize
is tuned by running 
on a subset
containing $20\%$ training data 
for a few epochs
(for the proposed method, we tune $\eta_1$).
All algorithms are implemented in Matlab.

%%%%%%%%%%%%%%%%%%%%%%%%%%%%%%%%%%%%%%%%%%%%%%%

\subsection{Strongly Convex Objectives}

\begin{figure}[h!t]
\begin{minipage}{.49\columnwidth}
\subfigure{\includegraphics[width=\columnwidth]{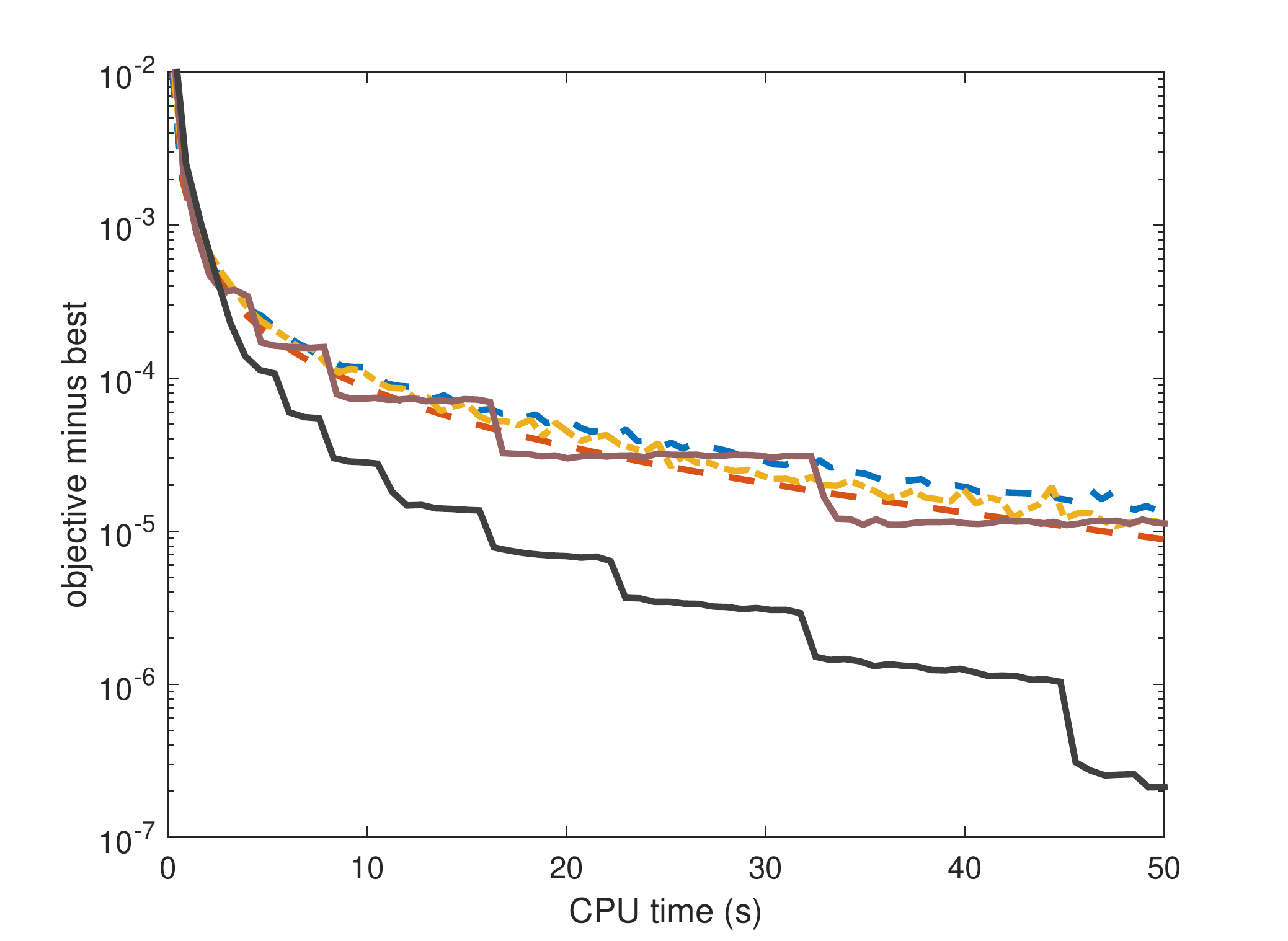}} \\
\vskip -.1in
\subfigure{\includegraphics[width=\columnwidth]{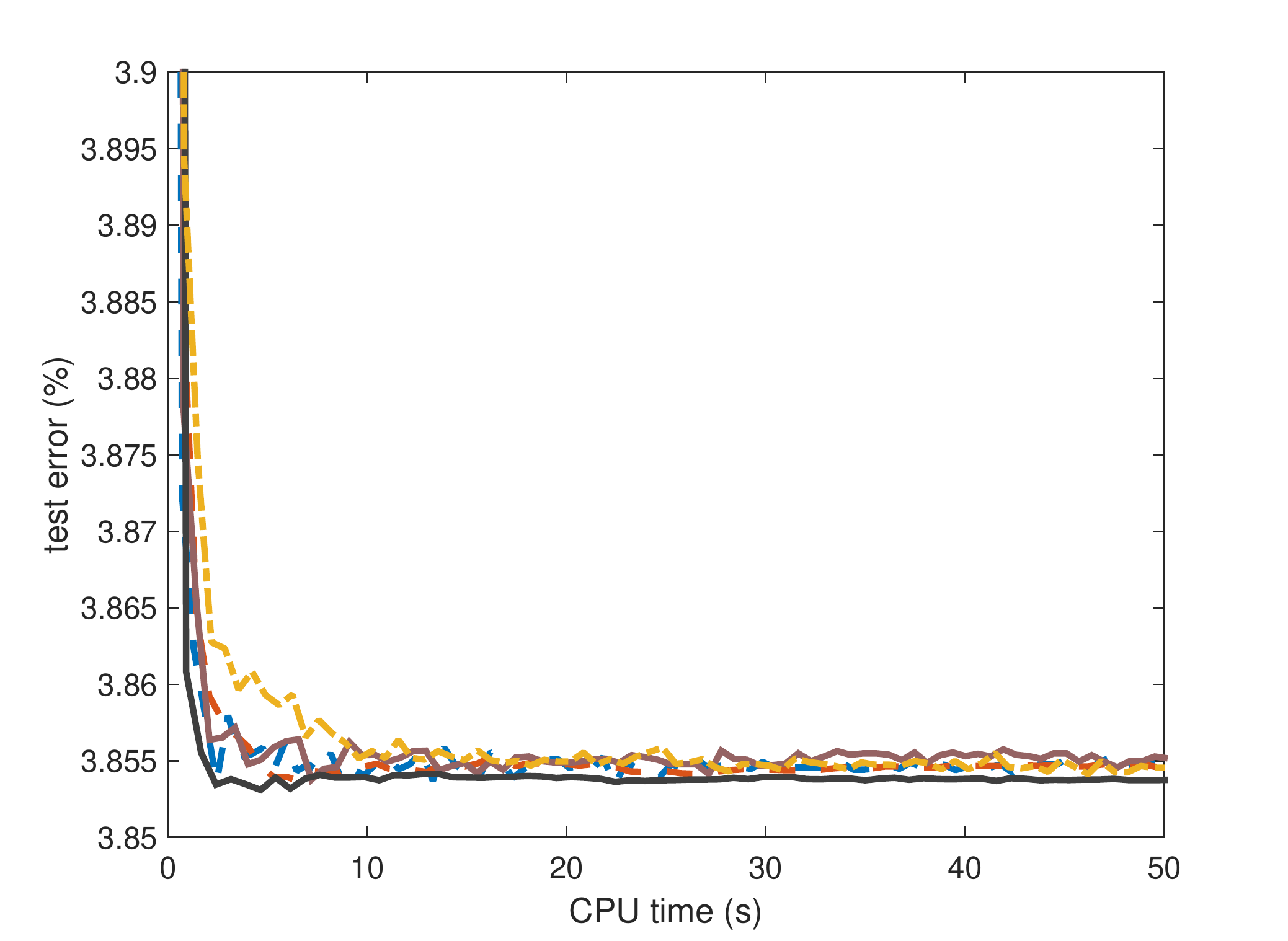}}
\centering
\text{\em{rcv1}.}
\vskip .1in
\end{minipage}
\begin{minipage}{.49\columnwidth}
\subfigure{\includegraphics[width=\columnwidth]{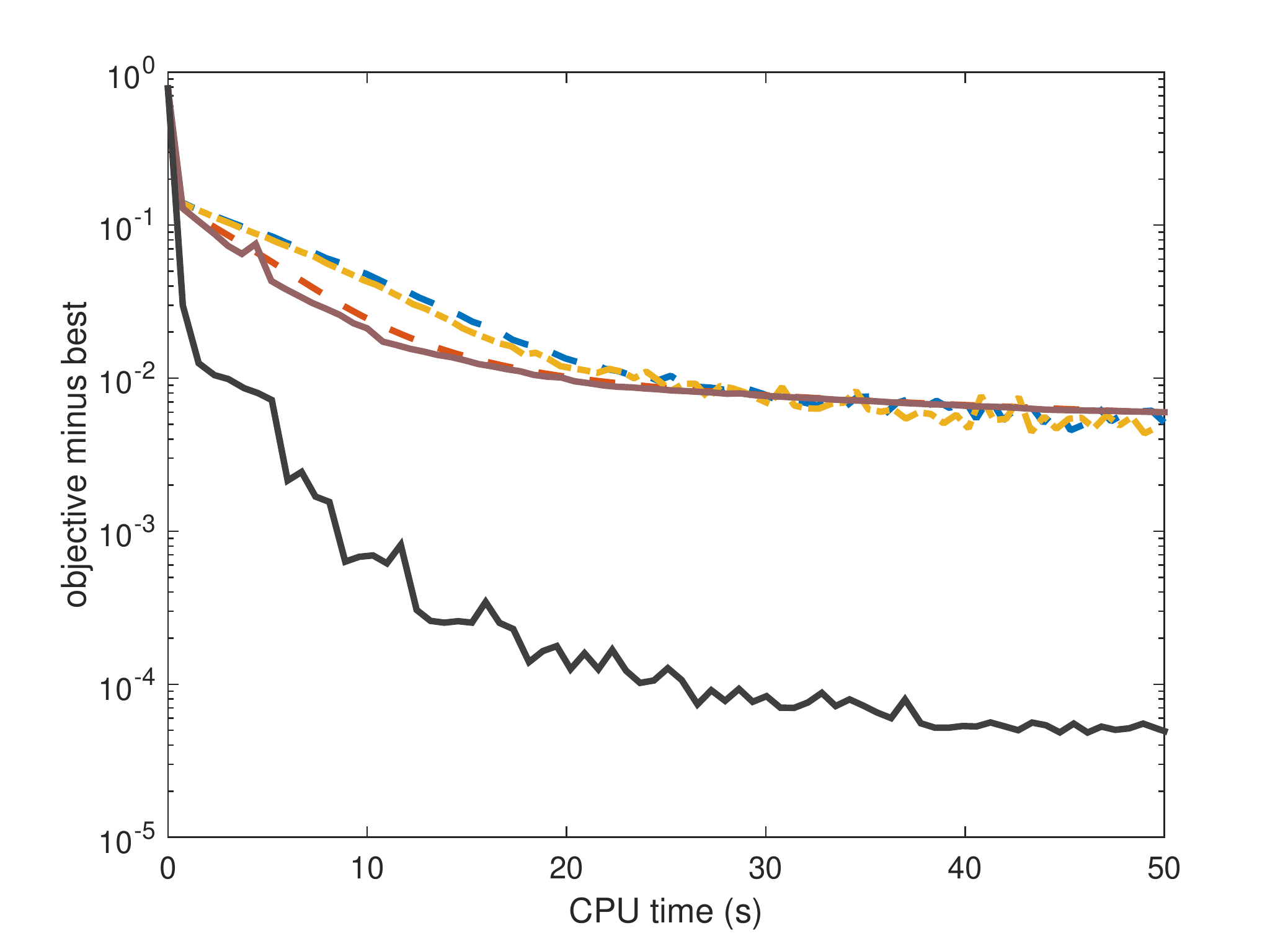}} \\
\vskip -.1in
\subfigure{\includegraphics[width=\columnwidth]{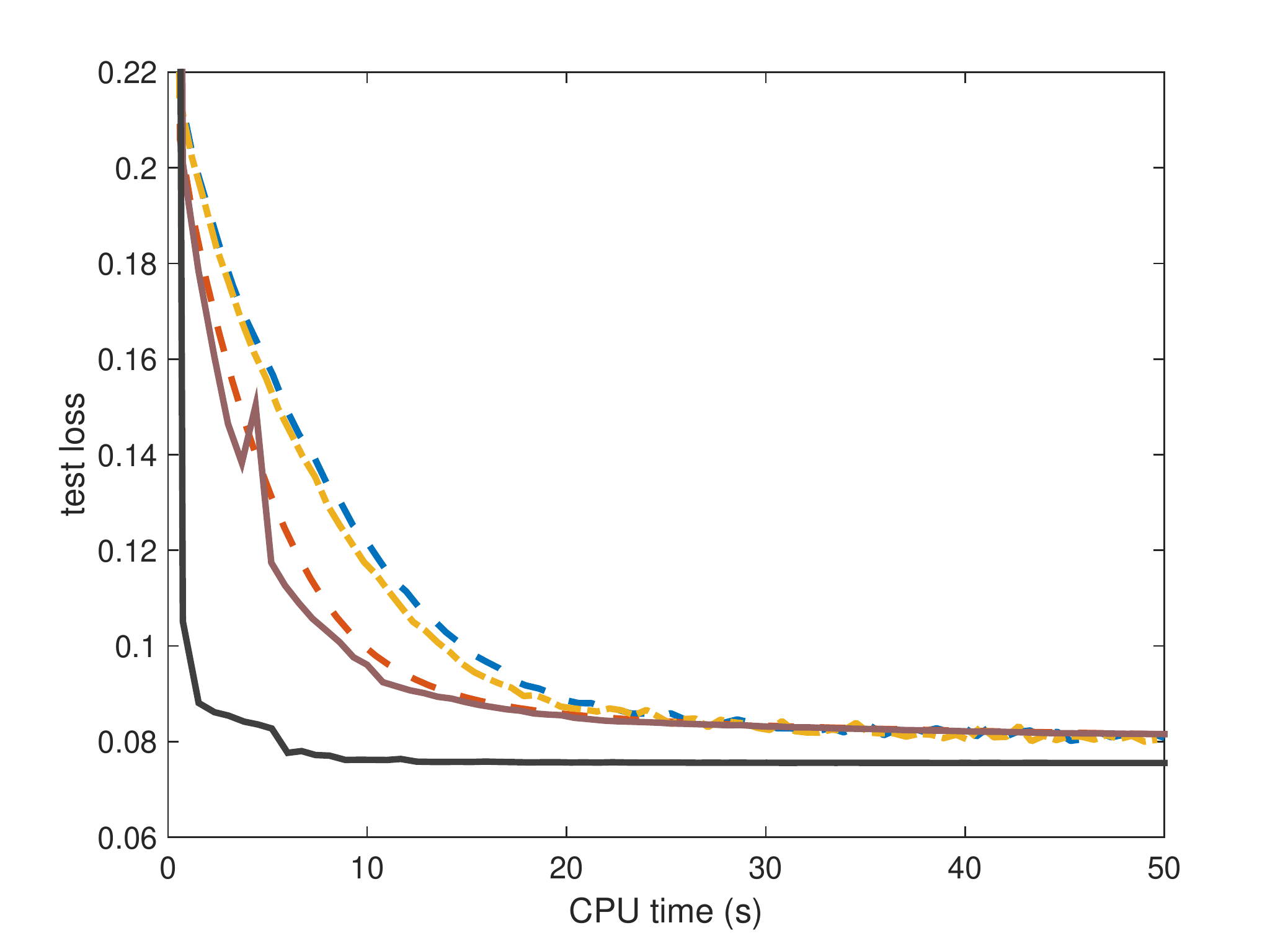}}
\centering
\text{\em{YearPredictionMSD}.}
\vskip .1in
\end{minipage}
\centerline{\includegraphics[width=2.5in,height=0.1in]{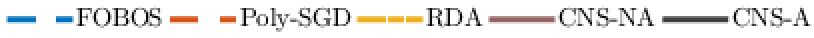}}
\vskip -.09in
\caption{Objective
(top) and
testing performance
(bottom)
vs CPU time (in seconds) on a strongly convex problem.}
\label{com_stoc_sp}
\end{figure}

Figure~\ref{com_stoc_sp} shows convergence of the objective and testing performance 
(classification error for \text{\em{rcv1}} and $\ell_1$-loss for \text{\em{YearPredictionMSD}}). 
The trends are consistent with Theorem~\ref{theorem:non-acc}.
CNS-A is the fastest (with a 
of
$O(1/T^2)$). This is followed by
CNS-NA and
Poly-SGD, both with $O(1/T)$ rate (from 
Theorem~\ref{theorem:non-acc} and
\cite{shamir2013stochastic}). The slowest are
FOBOS and RDA, which converge at a rate of $O(\log T/T)$ \cite{duchi2009efficient,xiao2009dual}.

\begin{figure}[h!]
\subfigure[{\em rcv1}.]{\includegraphics[width=.49\columnwidth]{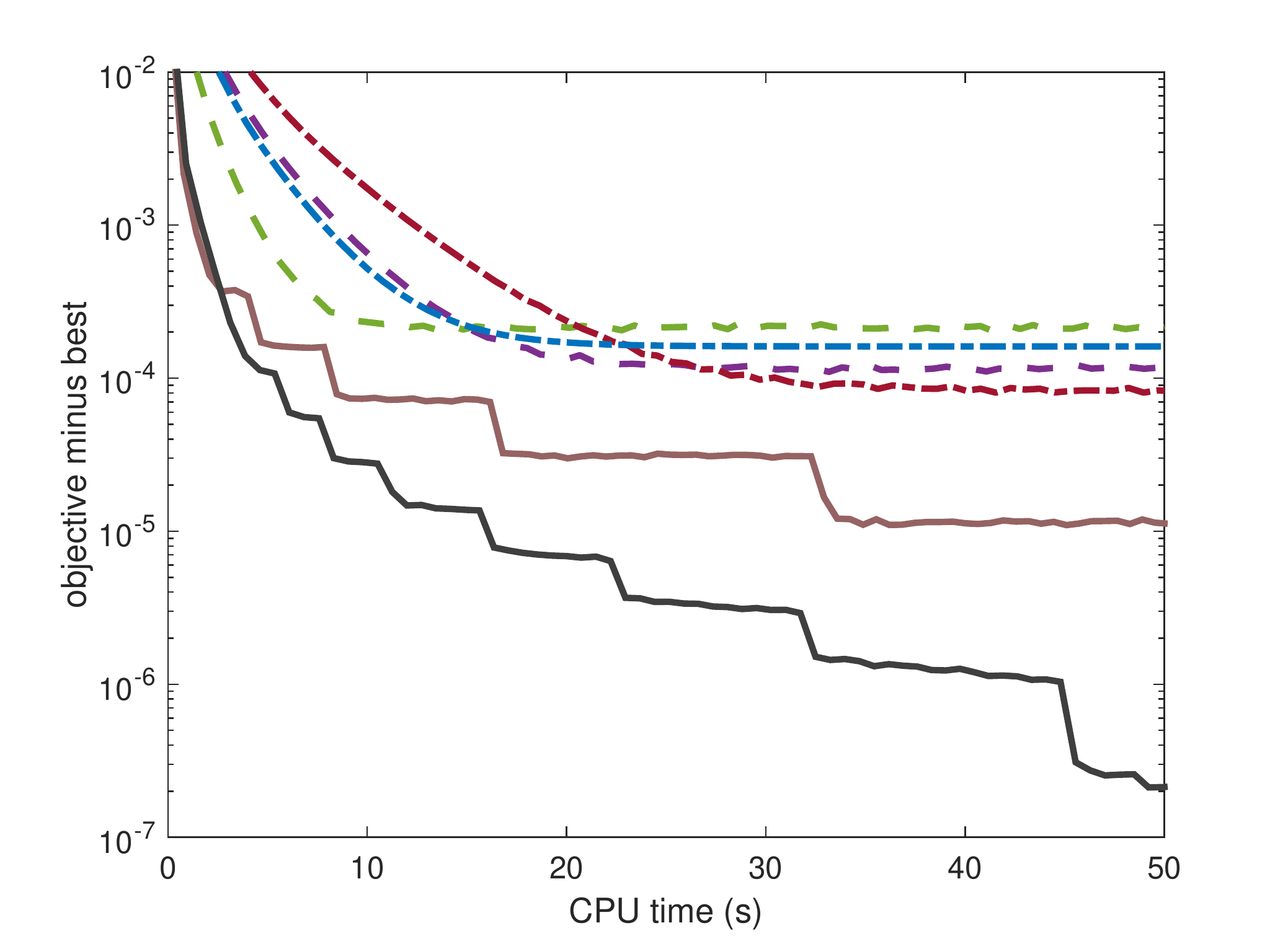}} 
\subfigure[{\em YearPredictionMSD}.]{\includegraphics[width=.49\columnwidth]{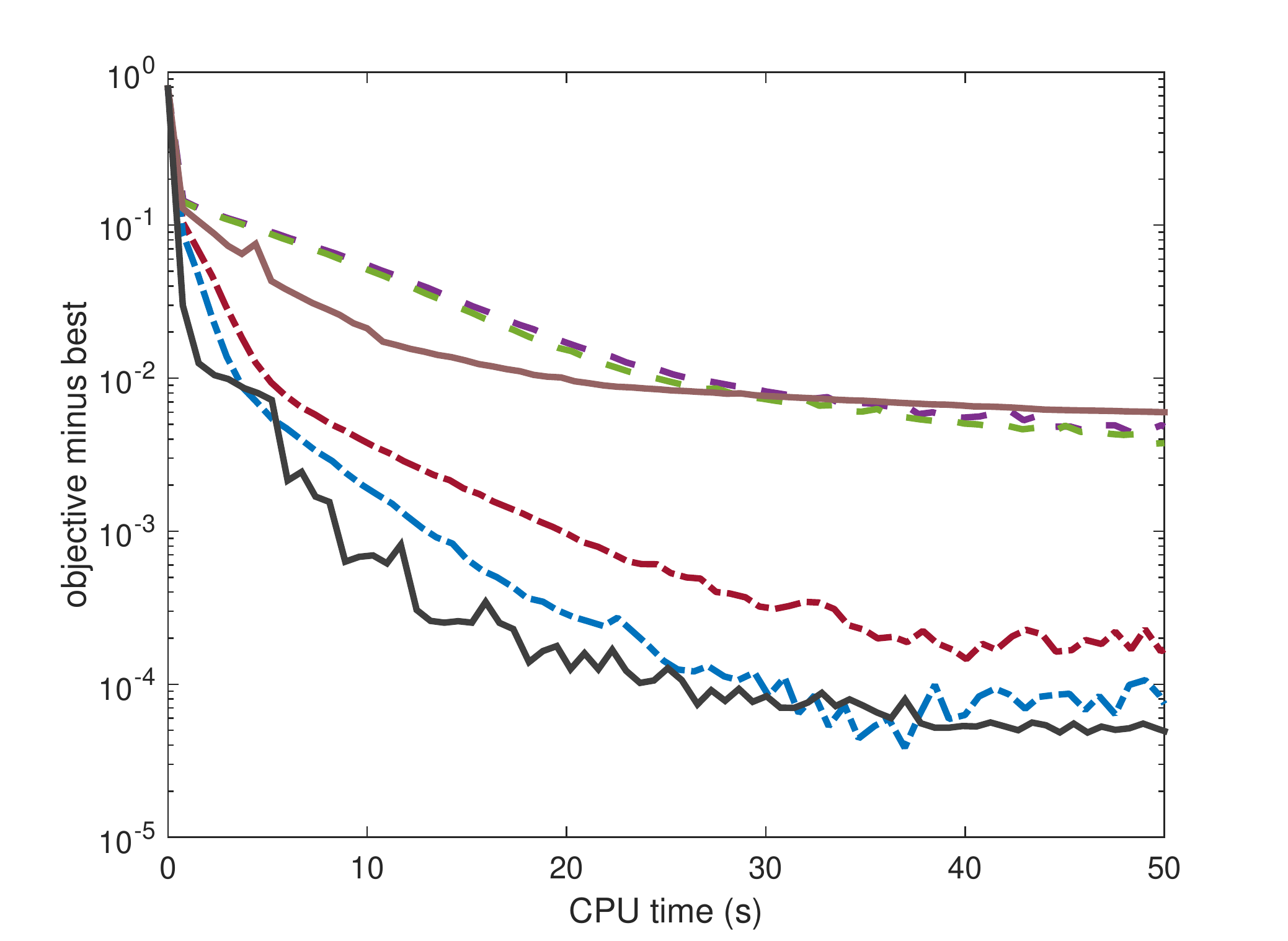}}
\centerline{\includegraphics[width=3.24in,height=0.12in]{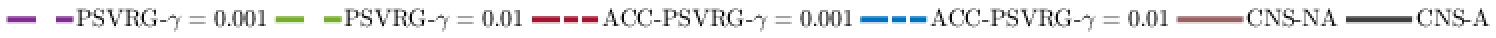}}
\vskip -.09in
\caption{Effect of continuation (strongly convex problem). 
}
\label{com_sm}
\end{figure}

Figure~\ref{com_sm} compares with the case where continuation is not used. 
Two fixed smoothness settings,
$\gamma =10^{-2}$ and $\gamma=10^{-3}$,
are used.
As can be seen, they are much slower (Propositions~\ref{prop:nonacc} and
\ref{prop:nonacc2}). Moreover,
a smaller $\gamma$ leads to slower convergence but better solution, while
a larger $\gamma$ leads to faster convergence but worse solution.
This is also consistent with Proposition~\ref{prop_rate_fixed}, as
using a fixed $\gamma$ only allows convergence to the optimal solution with a tolerance of $(1 +
\rho^S)\gamma D_u$.
Moreover, a
smaller
$\gamma$ leads to 
a larger
condition number, and convergence becomes slower.

%%%%%%%%%%%%%%%%%%%%%%%%%%%%%%%%%%%%%%%%%%

\subsection{General Convex Objectives}

\begin{figure}[h!]
\begin{minipage}{.49\columnwidth}
\subfigure{\includegraphics[width=\columnwidth]{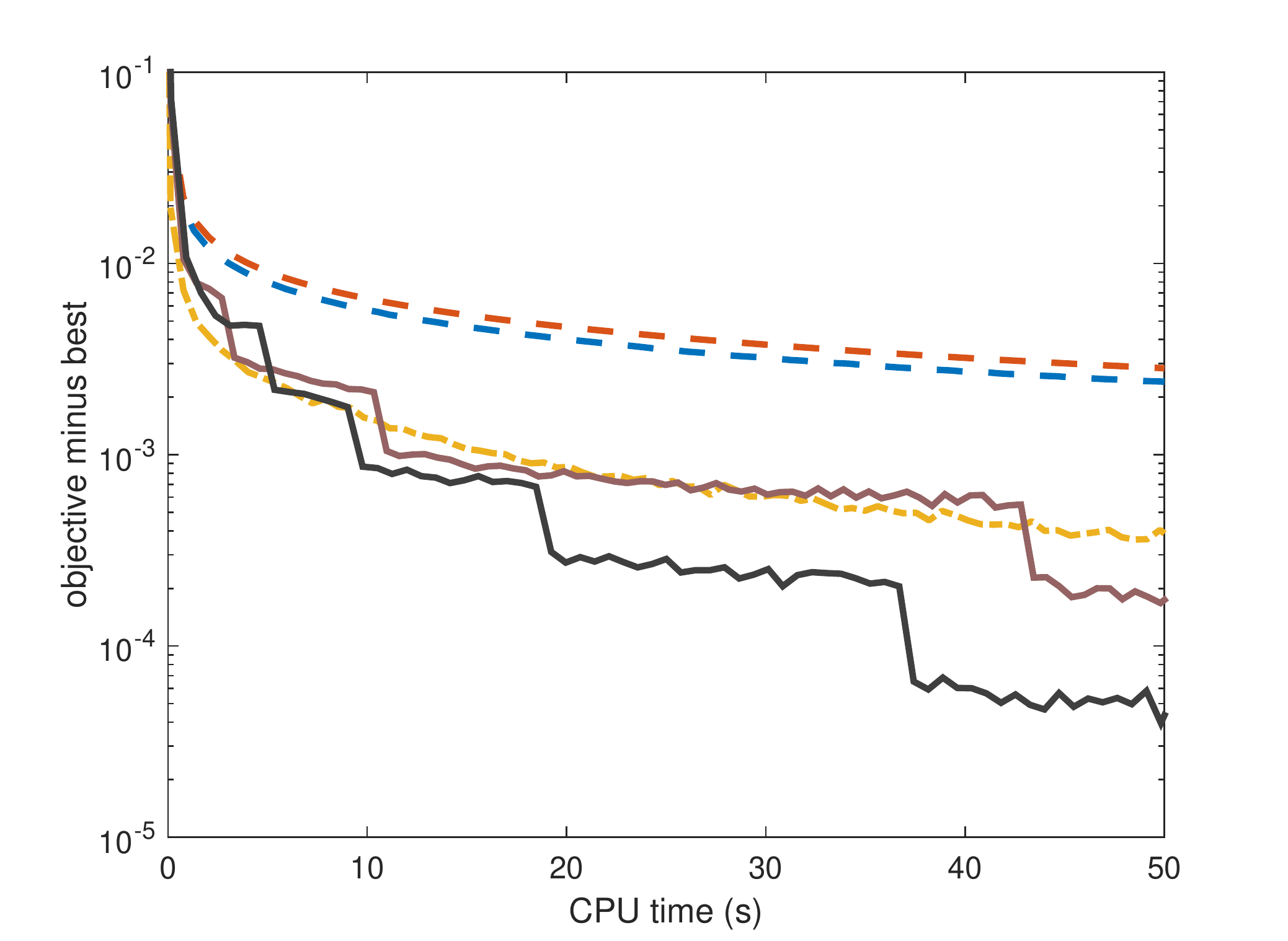}} \\
\vskip -.1in
\subfigure{\includegraphics[width=\columnwidth]{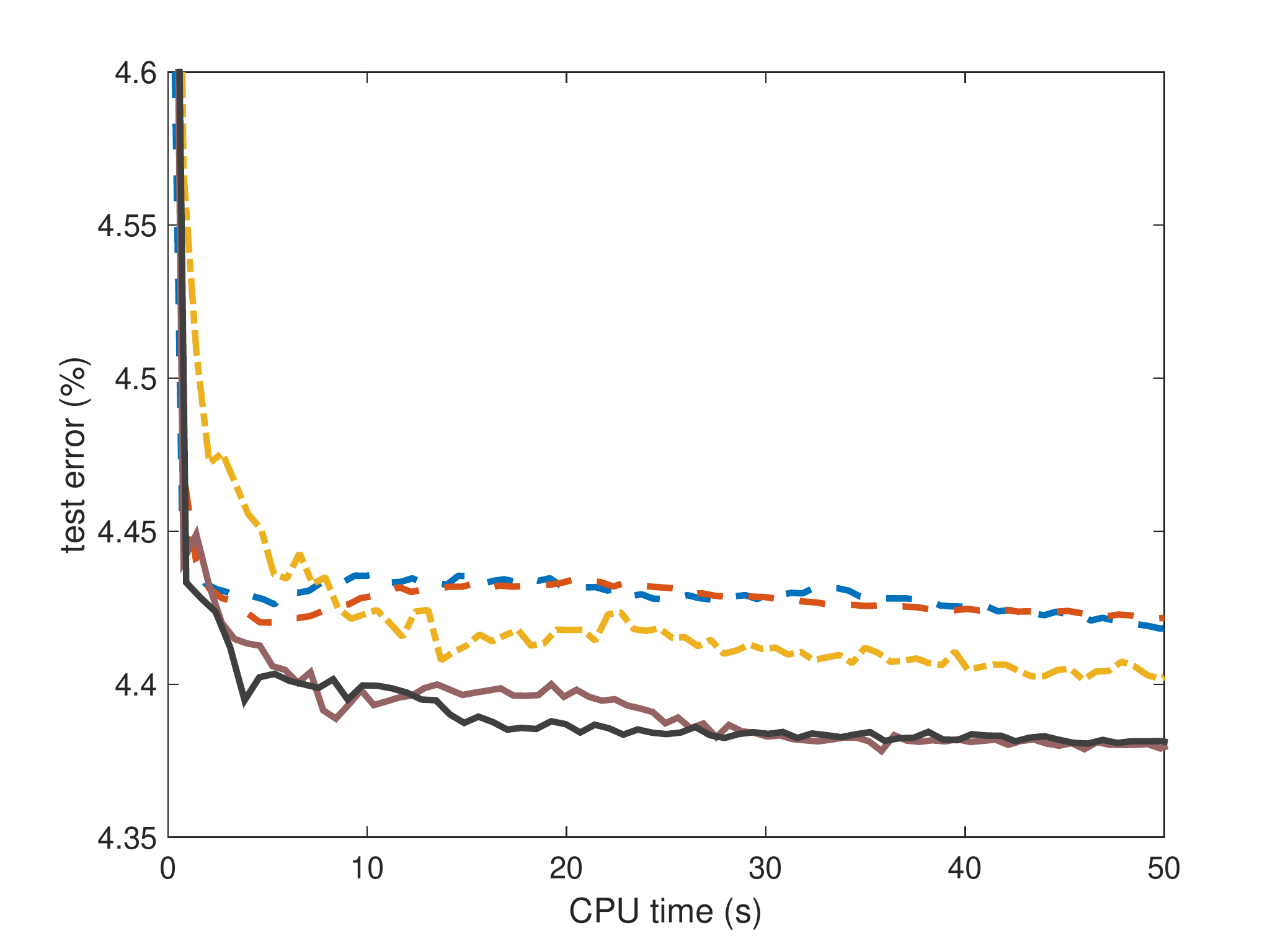}}
\centering
\text{\em{rcv1}.}
\vskip .1in
\end{minipage}
\begin{minipage}{.49\columnwidth}
\subfigure{\includegraphics[width=\columnwidth]{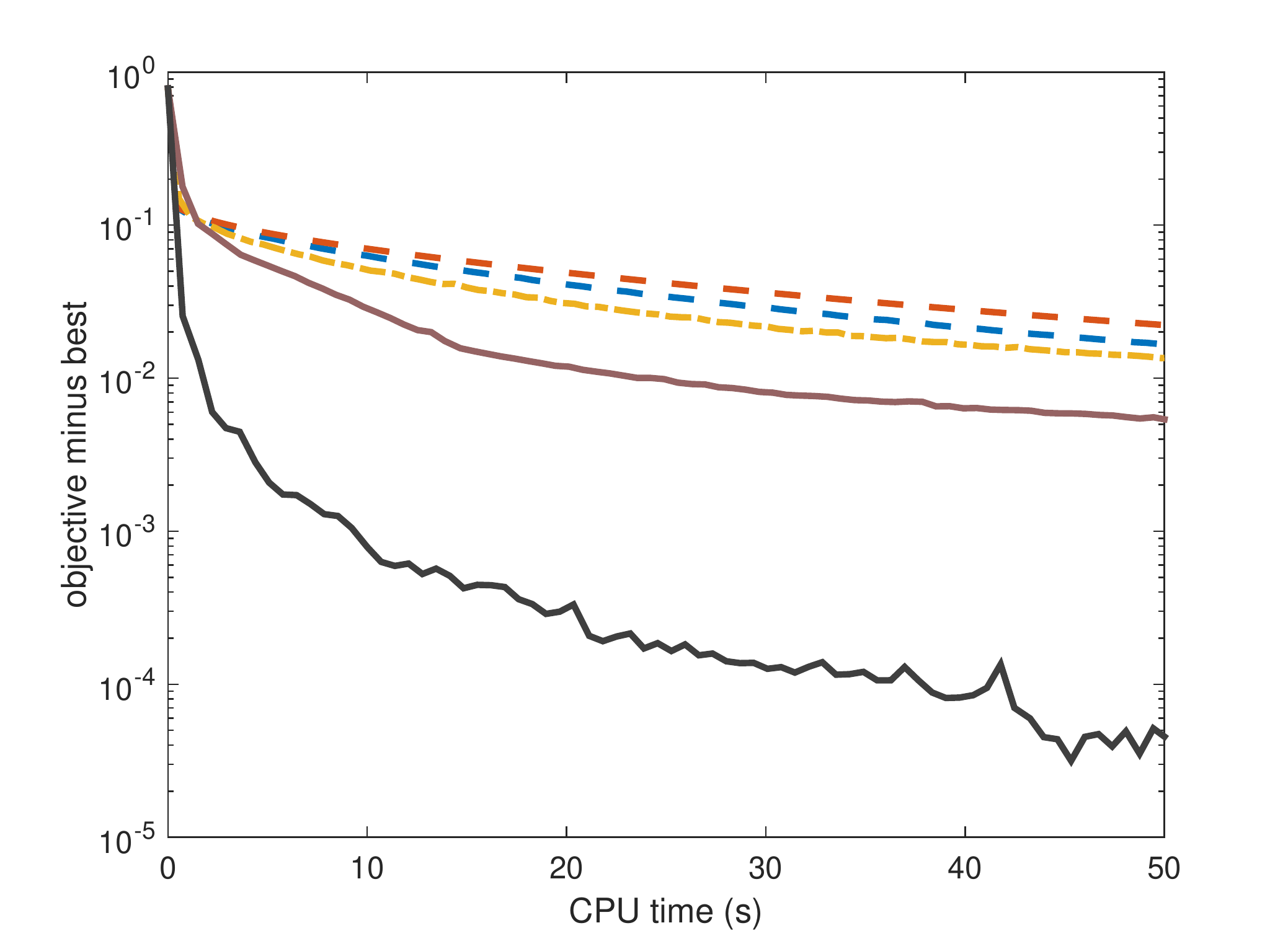}} \\
\vskip -.1in
\subfigure{\includegraphics[width=\columnwidth]{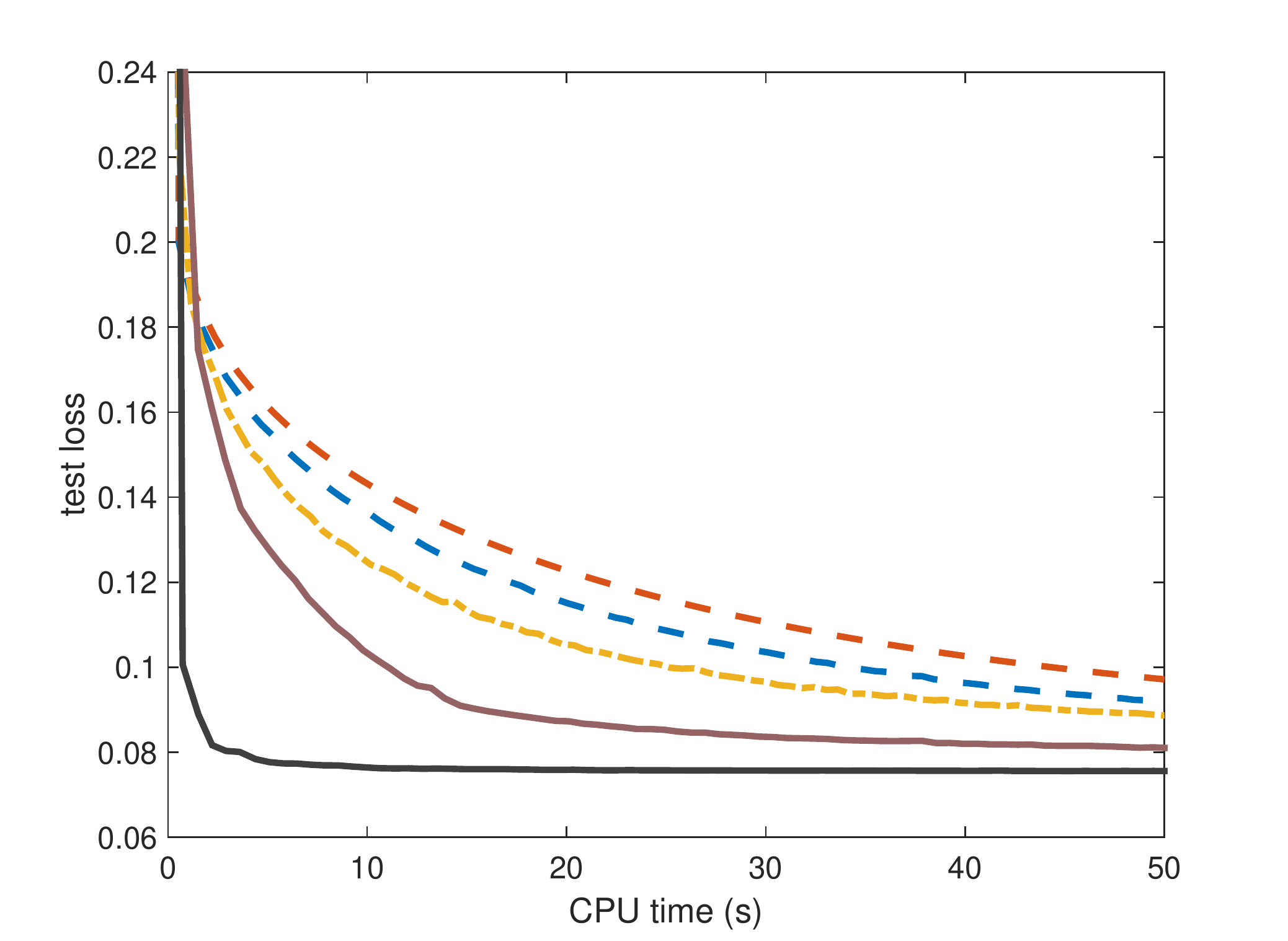}}
\centering
\text{\em{YearPredictionMSD}.}
\vskip .1in
\end{minipage}
\centerline{\includegraphics[width=2.5in,height=0.1in]{legend_sp}}
\vskip -.09in
\caption{Objective
(top)
and testing performance
(bottom)
vs CPU time (in seconds) on a general convex problem. 
}
\label{com_stoc_ns}
\end{figure}

We set $\lambda_1$ in Algorithm~\ref{alg:nsmm_nst} to  $10^{-5}$ for {\em rcv1},  
and $10^{-7}$ for {\em YearPredictionMSD}. As can be seen from 
Figure~\ref{com_stoc_ns},
the trends are again consistent with Theorem~\ref{theorem:non-acc-ns}.
CNS-A is the fastest
($O(1/T)$ convergence rate), while the others all have a
rate 
of $O(1/\sqrt{T})$ 
\cite{duchi2009efficient,xiao2009dual,shamir2013stochastic}. Also, RDA shows better performance than FOBOS
and Poly-SGD. 
Recall that Poly-SGD outperforms FOBOS and RDA
on strongly convex problems. 
However, on general convex problems, Poly-SGD is the worst as its
rate is only as good as others, and it does not exploit the composite structure of the problem.

\begin{figure}[h!]
\subfigure[{\em rcv1}.]{\includegraphics[width=.49\columnwidth]{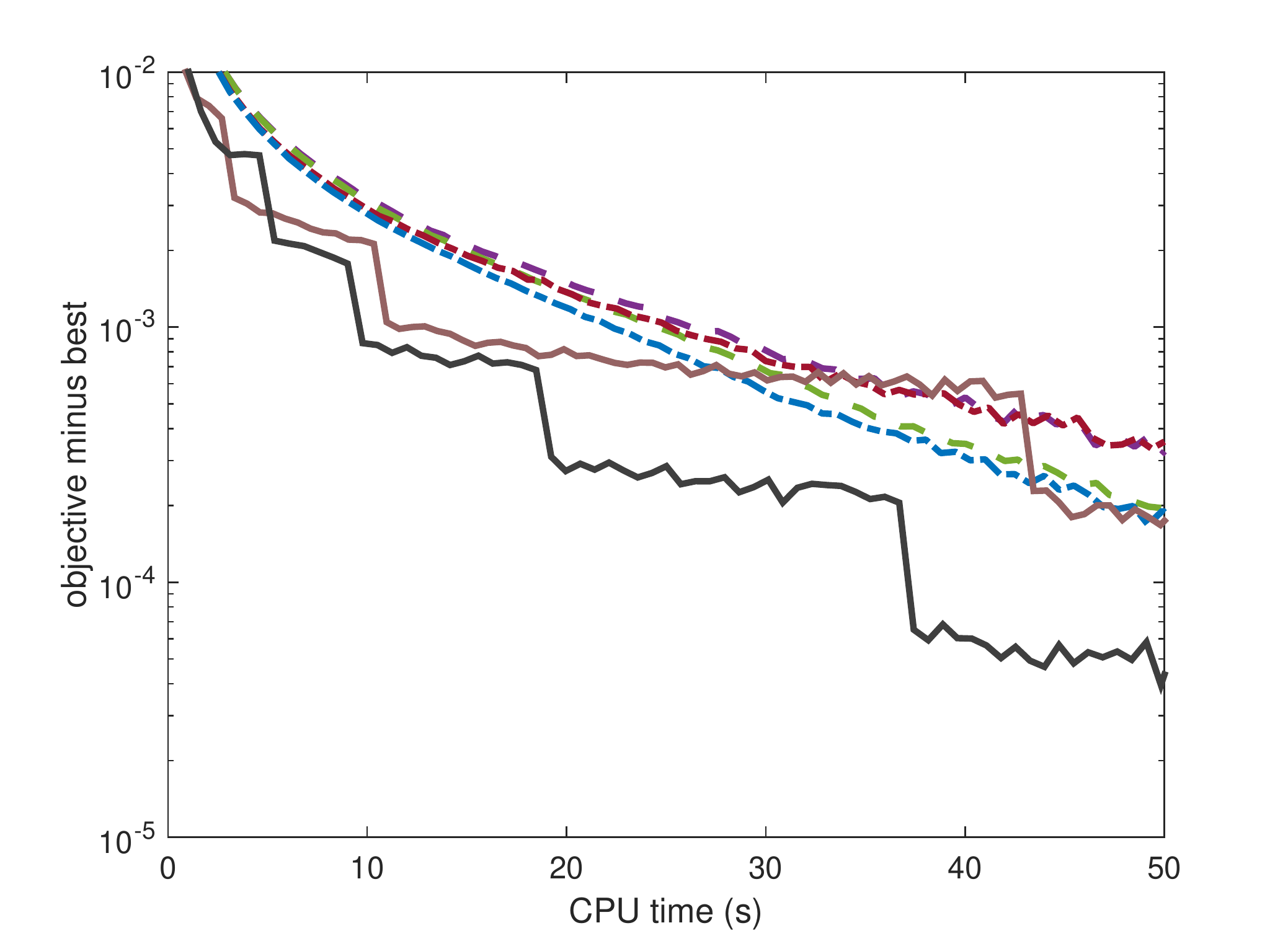}} 
\subfigure[{\em YearPredictionMSD}.]{\includegraphics[width=.49\columnwidth]{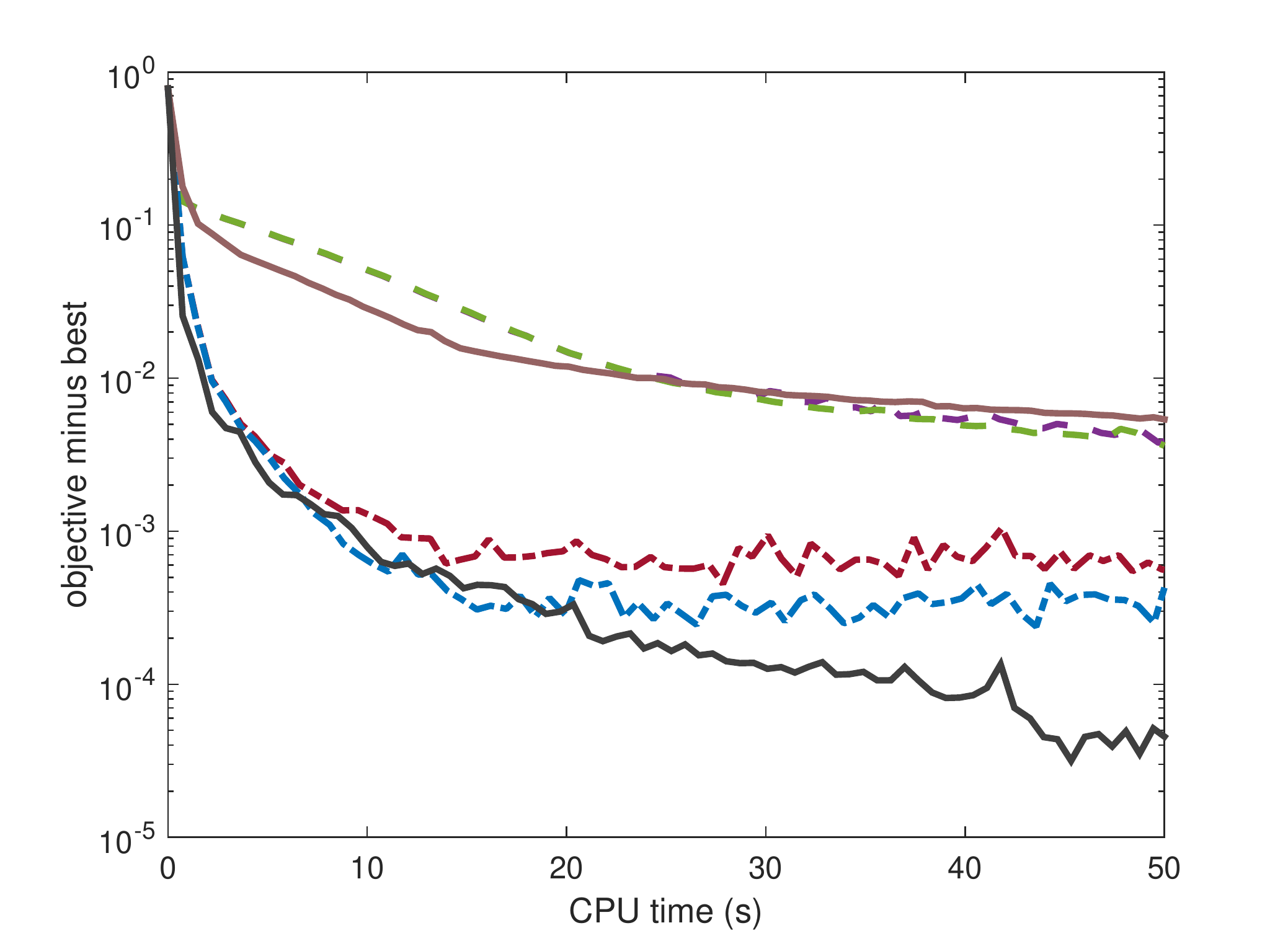}}
\vskip -.05in
\centerline{\includegraphics[width=3.24in,height=0.12in]{legend_sm}}
\vskip -.09in
\caption{Effect of continuation (general convex problem). 
}
\label{com_ns_sm}
\end{figure}

Figure~\ref{com_ns_sm} compares with the case where continuation is not used.
As in the previous section, CNS-NA and CNS-A show faster convergence than its
fixed-smoothing
counterparts.

%%%%%%%%%%%%%%%%%%%%%%%%%%%%%%%%%%%%%%%%%%%%%%%%%%%%%%%%%%%%%%%%%%%%%%%%%%%%%%%%%%%%

\section{Conclusion}

In this paper, we proposed a continuation algorithm (CNS) for regularized risk minimization
problems, in which both the loss and regularizer may be nonsmooth.  In each of its stages,
the smoothed subproblem can be easily solved by either existing accelerated or
non-accelerated solvers.  Theoretical analysis establishes convergence results on the whole
continuation algorithm, not just one of its stages.  In particular, when accelerated solvers
are used, the proposed CNS algorithm achieves the rate of $O(1/T^2)$ on strongly convex
problems, and $O(1/T)$ on general convex problems.  These are the fastest known rates for
nonsmooth optimization. However, CNS is advantageous in that it allows the use of a
regularizer (unlike the fastest batch algorithm) and can exploit the composite structure of
the optimization problem (unlike the fastest stochastic algorithm).  Experiments on nonsmooth
classification and regression models demonstrate that CNS outperforms the state-of-the-art.

\newpage

\section{Acknowledgments}

This research was supported in part by
the Research Grants Council of the Hong Kong Special Administrative Region
(Grant 614513).

\bibliographystyle{aaai}
\bibliography{nsmm}
\end{document}